\documentclass[11pt]{article}

\usepackage{acl}

\usepackage{times}
\usepackage{latexsym}

\usepackage[T1]{fontenc}

\usepackage[utf8]{inputenc}

\usepackage{microtype}

\usepackage{inconsolata}

\usepackage{amsmath}
\usepackage{cleveref}
\usepackage{booktabs}
\usepackage{graphicx}
\usepackage{amssymb}
\usepackage{xcolor}
\usepackage{xspace}

\newcommand{\methodname}{\texttt{news-crawler-LM}\xspace}
\definecolor{good}{RGB}{0,140,0}
\definecolor{bad}{RGB}{180,0,0}

\usepackage{tikz}
\usetikzlibrary{positioning,fit,calc,decorations.pathreplacing}
\usepackage{todonotes}

\title{\methodname: \\ A Small Long-Context Model For High-Quality News Crawling}

\author{Pascal Stolzenburg\thanks{Equal contribution} \quad Jonas Golde\footnotemark[1] \quad Max Dallabetta \quad Alan Akbik \\
  Humboldt-Universität zu Berlin \\}

\begin{document}
\maketitle
\begin{abstract}
Extracting structured content from news pages remains challenging due to heterogeneous HTML layouts, inconsistent markup, and substantial boilerplate such as navigation elements and advertisements. Rule-based news crawlers can achieve high extraction accuracy by encoding site-specific structure, but require manual configuration in order to generalize to new publishers. Large language models provide a more flexible alternative by reducing the need for handcrafted rules, but their high computational cost limits practical deployment. In this paper, we introduce \methodname, a small long-context language model fine-tuned on high-quality, human-validated extractions from the Fundus news-crawling library. Our model converts raw HTML into plaintext and structured JSON, including fields such as headline, author, publication date, and article body. In our experiments, \methodname outperforms strong baselines in HTML-to-Markdown and HTML-to-JSON extraction, improving performance by +4.8 BLEU and +6.1 METEOR in the HTML-to-Markdown task, and by +2.2 BLEU and +4.1 METEOR in the HTML-to-JSON task. However, we also observe that our model only slightly better compared to other rule-based parsing libraries on the HTML-to-plaintext task in evaluations on previously unseen publishers. We release all models and artifacts to the research community \footnote{\url{https://huggingface.co/stolzenp/FundusCrawler-plaintext}, \\ \url{https://huggingface.co/stolzenp/FundusCrawler-json}, \\ \url{https://huggingface.co/datasets/stolzenp/fundus-cleaned-filtered-62K}}.
\end{abstract}

\begin{table}[!htbp]
\centering
\setlength{\tabcolsep}{6pt}
\begin{tabular}{lccc}
\toprule
 & \textsc{Rules} & \textsc{ML} & \textsc{Ours} \\
\midrule
Scalability          & \textcolor{bad}{\texttimes}  & \textcolor{good}{\checkmark} & \textcolor{good}{\checkmark} \\
Extraction quality   & \textcolor{good}{\checkmark} & \textcolor{bad}{\texttimes}  & \textcolor{good}{\checkmark} \\
Cross-domain & \textcolor{bad}{\texttimes}  & \textcolor{good}{\checkmark} & \textcolor{good}{\checkmark} \\
Transparency         & \textcolor{good}{\checkmark} & \textcolor{bad}{\texttimes}  & \textcolor{good}{\checkmark} \\
\bottomrule
\end{tabular}
\caption{Comparison of news crawling approaches. \methodname (Ours) combines the strengths of rule-based (e.g., Fundus) and ML-based (e.g., LLMs) methods.}
\label{table:news-crawling-comparison}
\end{table}

\section{Introduction}
Online news articles are a widely used data source for training and evaluating language models, particularly for tasks such as social and political analysis \citep{masud2020hatenewinfodemictopicaware}, information extraction \citep{yates-etal-2007-textrunner,whitehouse-etal-2023-webie}, or document-level reasoning such as fact checking \citep{mishra-etal-2020-generating,sathe-etal-2020-automated}. However, training language models on web-derived news data requires converting raw HTML pages into clean, structured representations \citep{Hamborg2017}. This preprocessing step remains challenging due to heterogeneous page layouts, inconsistent markup, and substantial boilerplate content such as navigation elements and advertisements, which can introduce noise and bias into downstream model training \citep{penedo2024finewebdatasetsdecantingweb}.


\begin{figure*}[t!]
\centering
\begin{tikzpicture}[
    node distance=2mm,
    every node/.style={anchor=north west},
    html/.style={
        font=\fontsize{8}{8}\ttfamily\selectfont,
        text width=6cm,
        align=left
    },
    json/.style={
        font=\fontsize{8}{8}\ttfamily\selectfont,
        text width=5.2cm,
        align=left
    },
    box/.style={
        draw=black!50,
        rounded corners=8pt,
        inner sep=3pt
    },
]

\node[html] (html) {
\textcolor{black!60}{<html>}\\
\ \ \textcolor{black!60}{<body>}\\
\ \ \ \textcolor{black!60}{<article>}\\
\ \ \ \ \textcolor{black!60}{<h1>}Oppenheimer wins Oscars\textcolor{black!60}{</h1>}\\
\ \ \ \ \textcolor{black!60}{<span class="author">}by Max Mustermann\textcolor{black!60}{</span>}\\
\ \ \ \ \textcolor{black!60}{<p class="summary">}\\
\ \ \ \ \ \ Oppenheimer wins 7 Oscars in 2024...\\
\ \ \ \ \textcolor{black!60}{</p>}\\
\ \ \ \ \textcolor{black!60}{<p>}Christopher Nolan’s blockbuster...\\
\ \ \ \ \textcolor{black!60}{</p>}\\
\ \ \ \ \textcolor{black!60}{<h2>}Nolan's first Oscar\textcolor{black!60}{</h2>}\\
\ \ \ \ \textcolor{black!60}{<p>}This marks director Christopher Nolan's first...\\
\ \ \ \ \textcolor{black!60}{</p>}\\
\ \ \ \ \textcolor{black!60}{<ul class="tags">}\\
\ \ \ \ \ \ \textcolor{black!60}{<li>}Entertainment\textcolor{black!60}{</li>}\\
\ \ \ \ \ \ \textcolor{black!60}{<li>}Oscars 2024\textcolor{black!60}{</li>}\\
\ \ \ \ \textcolor{black!60}{</ul>}\\
\ \ \ \textcolor{black!60}{</article>}\\
\ \ \textcolor{black!60}{</body>}\\
\textcolor{black!60}{</html>}
};

\node[box, fit=(html)] (htmlbox) {};

\node[font=\fontsize{7}{7}\bfseries\selectfont, anchor=south west]
    at (htmlbox.north west) {HTML input};

\node (arrowstart) [right=9mm of htmlbox.east, anchor=west] {};
\node (arrowend)   [right=9mm of arrowstart.east, anchor=west] {};

\draw[->, very thick]
    (arrowstart) -- (arrowend)
    node[
        midway,
        above,
        yshift=2.5mm,
        draw=blue!60,
        fill=blue!10,
        rounded corners=3pt,
        inner sep=2pt,
        font=\fontsize{10}{10}\selectfont
    ]
    {\methodname};

\node[json, right=35mm of htmlbox.east] (json) {
\{\\
\ \ "title": "Oppenheimer wins Oscars",\\
\ \ "authors": ["Max Mustermann"],\\
\ \ "body": \{\\
\ \ \ \ "summary": "Oppenheimer wins 7 Oscars in 2024...",\\
\ \ \ \ "paragraphs": [\\
\ \ \ \ \ \ "Christopher Nolan’s blockbuster...",\\
\ \ \ \ \ \ "This marks director Christopher Nolan's first..."\\
\ \ \ \ ],\\
\ \ \ \ "subheadlines": ["Nolan's first Oscar"]\\
\ \ \},\\
\ \ "topics": ["Entertainment", "Oscars 2024"],\\
\ \ "plaintext": "Oppenheimer wins 7 Oscars in 2024... Christopher Nolan’s blockbuster..."\\
\}
};

\node[box, fit=(json)] (jsonbox) {};

\node[font=\fontsize{7}{7}\bfseries\selectfont, anchor=south west]
    at (jsonbox.north west) {Structured output / plaintext};

\end{tikzpicture}

\caption{
HTML conversion task: Given the raw HTML of a crawled news article, the model is trained to extract the article content (HTML-to-plaintext) and structured attributes, e.g.\ content, title, authors, topics (HTML-to-JSON).
}
\vspace{-2mm}
\label{fig:attribute_overview}
\end{figure*}
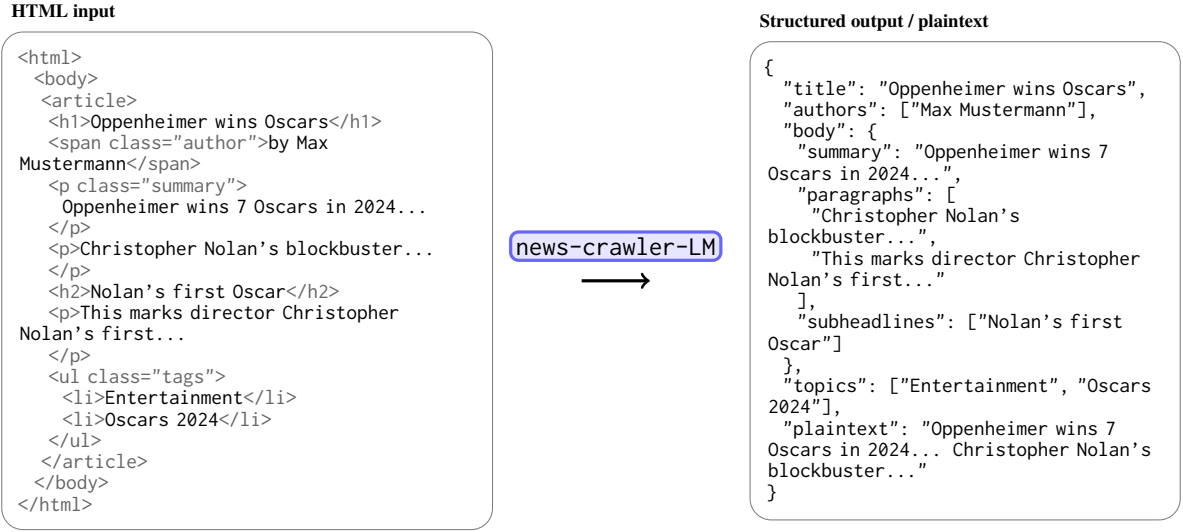

Current approaches for such conversion tasks show complementary limitations. First, rule-based extraction systems such as \textsc{Fundus} \citep{dallabetta-etal-2024-fundus} or Trafilatura \citep{barbaresi-2021-trafilatura} encode explicit assumptions about website structure and can produce high-quality outputs, but require manual rule-engineering and frequent updates which makes it challenging to scale this approach to a large and evolving set of publishers. On the other end, large language models (LLMs) can be applied with minimal configuration and generalize across diverse HTML layouts, but their high computational cost and limited precision in structured extraction constrain their use for large-scale data preparation \citep{gur-etal-2023-understanding,shen-etal-2025-autoclean}. More fundamentally, we hypothesize the HTML-conversion task may be most effective using small and efficient language models trained on low-noise, human-validated HTML-to-text pairs.

To this end, we present \methodname, a small long-context language model designed to bridge the gap between rule-based and generative extraction approaches. Our approach combines three components: \textit{(i)} human-validated training data derived from the Fundus library, resulting in low-noise supervision; \textit{(ii)} a strong Transformer backbone, namely ReaderLM-v2 \citep{wang2025readerlmv2smalllanguagemodel}, which is pre-trained for the HTML-to-text conversion task; and \textit{(iii)} a contrastive training objective to reducing degeneration at inference time when processing long HTML sequences. Specifically, we train our model to perform two tasks: \textit{(i)} HTML-to-plaintext, and \textit{(ii)} HTML-to-JSON, containing fields such as title, author, publication date, and article body. In our evaluations, we find that \methodname outperforms competitive long-context baselines, including ReaderLM-v2 and Qwen2.5-32B. However, we also observe that performance is comparable to other parsing libraries beyond Fundus on the HTML-to-plaintext task, suggesting that for simpler settings, machine learning–based approaches may not be necessary to generalize to previously unseen publisher sites.

We summarize our contributions as follows:
\begin{enumerate}
    \item We introduce \methodname, a small long-context language model for HTML-to-plaintext and HTML-to-JSON task for news articles.
    \item We empirically evaluate \methodname with rule-based systems and strong long-context language models in zero-shot publisher settings.
    \item We release model checkpoints, datasets, and training scripts to support reproducible research and practical deployment.\footnote{Removed for double-blinded review.}
\end{enumerate}

\section{\methodname}

\subsection{Dataset}

\noindent\textbf{Desiderata.}~
Prior work on HTML conversion typically relies on large-scale crawling and heuristic filtering to construct training data. For example, ReaderLM-v2 obtains one million articles from CommonCrawl, where heuristics are applied to derive corresponding markdown representations. In contrast, we sample approximately 100k HTML-to-plaintext and JSON pairs extracted using human-defined rules from the Fundus library, and investigate whether this low-noise, high-quality supervision is sufficient to generalize across publishers.

\noindent\textbf{Crawling Data From Fundus.}~We construct the training data $\mathcal{D}$ using the Fundus library \citep{dallabetta-etal-2024-fundus}. Formally, we create the training data as a set of input–output pairs
\(
\mathcal{D} = \{(x_i, y_i)\}_{i=1}^{N},
\)
where each input \(x_i\) represents a cleaned HTML of a news article and each target \(y_i\) represents the extraction output. Depending on the task setting, \(y_i\) is either plaintext or a structured JSON serialization containing fields such as title, author, publication date, and article body.

At the time of writing, we consider all 106 publishers supported by the Fundus library. For each publisher, we collect up to 1,000 articles to capture recurring publisher-specific HTML patterns. During crawling, 28 of the 106 publishers could not be fully retrieved due to rate limitations or changes in HTML structure. When possible, we supplement missing articles with older content from the CC-News dataset, which is natively supported by Fundus. After this process, we obtain a complete set of articles for 93 publishers, which constitute our core dataset spanning 25 languages.


\noindent\textbf{Preprocessing and Filtering.}~We aim to support a context window of 32k tokens, reserving 24k tokens for the HTML input and 8k tokens for output generation. We filter irrelevant content such as JavaScript and \texttt{<img>}-tags following \citet{wang2025readerlmv2smalllanguagemodel}, and discard documents whose HTML input exceeds the 24k token limit after filtering. This retains $\sim$79\% of the data. None of the outputs (plaintext or JSON) exceeds the 8k token limit. We show the subword token distributions for HTML (input), plaintext, and JSON (targets) in \Cref{fig:token_distribution}.



\subsection{Model}
\noindent\textbf{Desiderata.}~
General-purpose LLMs demonstrate strong zero-shot generalization and can handle HTML conversion without any task-specific training. However, their high computational cost makes them impractical for large-scale crawling pipelines. We therefore aim to distill this capability into a small, task-specialized model that approximates the behavior of human-written extraction rules while remaining efficient and scalable.

\noindent\textbf{Backbone Model.}~
We choose ReaderLM-v2 as our transformer backbone which has been continually pre-trained on HTML and markdown texts and post-trained using supervised fine-tuning (SFT) and direct preference optimization (DPO) \citep{rafailov2024directpreferenceoptimizationlanguage} on HTML-to-markdown from randomly sampled CommonCrawl website\footnote{\url{https://commoncrawl.org/blog/common-crawl-url-index}}. ReaderLM-v2 builds upon Qwen2.5 \citep{qwen2025qwen25technicalreport} and thus supports a context length of up to 32k tokens. Rather than aiming for diversity, we fine-tune the model on a small but high-quality dataset $\mathcal{D}$ using a standard language modeling objective, minimizing the negative log-likelihood for each training pair $(x_i, y_i)$:
\[
\mathcal{L} = - \sum_{t=1}^{T} \log p(y_{i,t} \mid y_{i,<t}, x_i).
\]

\subsection{Contrastive Learning For Isotopic Token Representations}
Language models trained with a cross-entropy objective are known to exhibit text degeneration, particularly for long-context generation \citep{jiang2022simplecontrastivelearningobjective}. To mitigate this issue, we augment the standard language modeling objective with a contrastive loss following SimCTG \citep{su2022contrastiveframeworkneuraltext}, which discourages overly similar token representations and reduces repetitive or unstable generation behavior.

Specifically, let $x = (x_1, \ldots, x_{|x|})$ denote the input sequence with corresponding hidden representations $(h_{x_1}, \ldots, h_{x_{|x|}})$. The contrastive term is defined using cosine similarity and a margin $\rho \in [-1,1]$:
\[
f(x_i, x_j) =
\max (
  0,\,
  \rho - s(h_{x_i}, h_{x_i}) + s(h_{x_i}, h_{x_j})
)
\]

\noindent with
\[
s(h_{x_i}, h_{x_j}) =
\frac{h_{x_i}^\top h_{x_j}}{\lVert h_{x_i} \rVert \, \lVert h_{x_j} \rVert}.
\]

\noindent Thus, the overall contrastive loss becomes:
\[
\mathcal{L}_{\text{CL}} =
\frac{1}{|x| (|x|-1)}
\sum_{i=1}^{|x|}
\sum_{\substack{j=1 \\ j \neq i}}^{|x|}
f(x_i, x_j),
\]
encouraging isotropic token representations. Following \citet{su2022contrastiveframeworkneuraltext}, we set $\rho = 0.5$. Our final training objective is:
\[
\mathcal{L} = \mathcal{L}_{\text{LM}} + \mathcal{L}_{\text{CL}}.
\]

\section{Experimental Setup}

\subsection{Data}

We split the dataset by publisher into training, validation, and test sets, ensuring that no publisher appears in more than one split. We reserve 7 publishers for validation ($\sim$7.7k samples) and sample 100 HTML inputs from each of 10 held-out publishers for testing, with the remaining data used for training.

We show the distribution of languages in the overall dataset in \Cref{tab:language-distribution} and observe that the majority of articles are in Latin script. However, there are also publishers using Japanese or Hindi script.

\begin{table}[h!]
\centering
\begin{tabular}{l r}
\toprule
\textsc{Language} & \textsc{Share (\%)} \\
\midrule
German      & 50.4 \\
English     & 36.5 \\
French      & 3.2 \\
Norwegian   & 3.2 \\
Turkish     & 2.2 \\
Spanish     & 1.2 \\
Hindi       & 1.1 \\
Japanese    & 1.1 \\
Others      & 1.1 \\
\bottomrule
\end{tabular}
\caption{Language Distribution of Crawled Articles}
\label{tab:language-distribution}
\end{table}

\begin{figure*}[ht]
    \centering
    \includegraphics[width=0.9\linewidth]{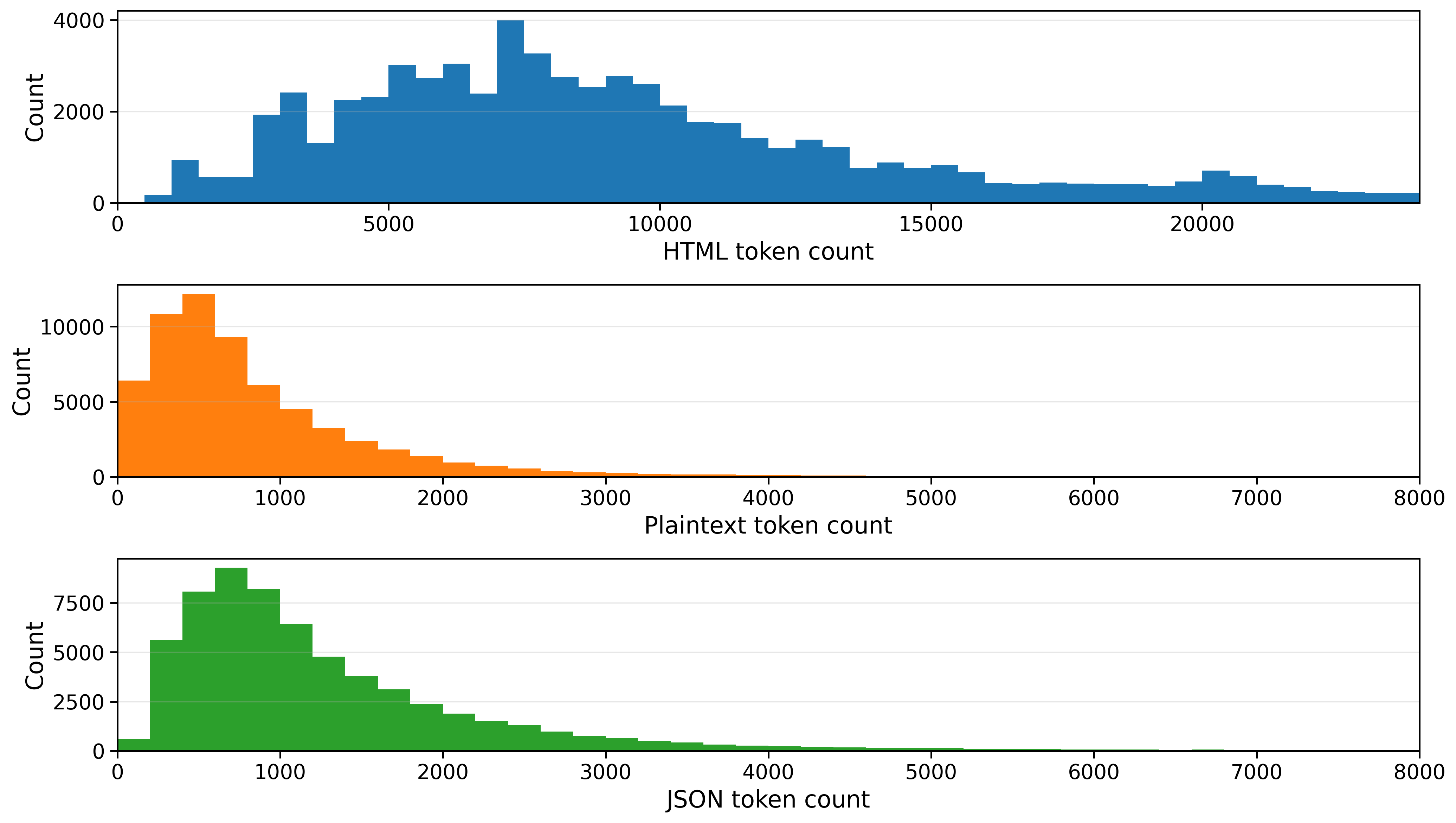}
    \caption{Distributions of subword token counts of the training data. We show HTML input counts (top), plaintext output counts (middle), and JSON output counts (bottom).}
    \label{fig:token_distribution}
\end{figure*}

Further, we show the distribution of subword token counts for HTML inputs, plaintext and JSON outputs in \Cref{fig:token_distribution}.

\subsection{Training}
We train the model using 8-bit Adam \citep{kingma2017adammethodstochasticoptimization, dettmers2022optimizers} with a batch size of $1$, a learning rate of $4e^{-5}$, and a weight decay of $0.01$. Training runs for 60k steps with a linear warmup over the first 1k steps, after which the learning rate decays linearly to zero. We select the best checkpoint based on validation perplexity with early stopping (patience $= 3$), evaluating every 1k steps. For parameter-efficient fine-tuning, we employ LoRA \citep{hu2021loralowrankadaptationlarge} with rank $r = 16$, scaling factor $\alpha = 16$, and no dropout, applying adapters to the attention projections ($q$, $k$, $v$, $o$) as well as the gating and feed-forward layers. This configuration keeps the number of trainable parameters small while allowing the model to adapt effectively to the HTML conversion task.

\begin{table*}[ht]
\centering
\begin{tabular}{lccccc}
\toprule
\textsc{Model} &
\textsc{BLEU} $\uparrow$ &
\textsc{METEOR} $\uparrow$ &
\textsc{ROUGE-L} $\uparrow$ &
\textsc{Lev.} $\downarrow$ &
\textsc{Jaro-W.} $\uparrow$ \\
\midrule
\textit{Parsing libraries} \\
news-please & 0.599 & \underline{0.702} & 0.669 & 0.386 & 0.741 \\
Boilerpipe & 0.599 & 0.654 & 0.680 & 0.380 & 0.728 \\
\midrule
\textit{Pre-trained language models} \\
ReaderLM-v2 & 0.098 & 0.294 & 0.275 & 0.830 & 0.544 \\
Qwen2.5-32B-Instruct & 0.077 & 0.249 & 0.168 & 0.821 & 0.601 \\
\midrule
\textit{Fine-tuned language models} \\
Qwen2.5-1.5B & 0.590 & 0.652 & 0.677 & 0.384 & 0.774 \\
Qwen2.5-1.5B + SimCTG & \underline{0.623} & 0.678 & \underline{0.707} & \underline{0.342} & \underline{0.786} \\
\methodname & \textbf{0.671} & \textbf{0.739} & \textbf{0.749} & \textbf{0.295} & \textbf{0.820} \\
\bottomrule
\end{tabular}
\caption{Results for HTML-to-plaintext task.}
\label{tab:plaintext-results}
\end{table*}

\subsection{Metrics} \label{sec:metrics}

\noindent\textbf{BLEU} \citep{papineni-etal-2002-bleu} measures n-gram precision between the predicted and reference text, combined with a brevity penalty to discourage overly short outputs. We compute BLEU using up to 4-grams.

\noindent\textbf{METEOR} \citep{banerjee-lavie-2005-meteor} aligns unigrams between prediction and reference, supporting exact, stemmed, and synonymous matches. Unlike BLEU, it accounts for word order by penalizing fragmented alignments. We use the default parameterization of \citet{banerjee-lavie-2005-meteor}.

\noindent\textbf{ROUGE-L} \citep{lin-2004-rouge} measures overlap via the longest common subsequence (LCS) between prediction and reference, capturing in-order matches without requiring contiguous spans.

\noindent\textbf{Levenshtein Distance} \citep{Levenshtein1965BinaryCC} counts the minimum number of character-level edit operations (insertions, deletions, substitutions) needed to transform the prediction into the reference. We report the normalized distance to account for varying sequence lengths.

\noindent\textbf{Jaro-Winkler Similarity} \citep{winkler1990string} extends the Jaro metric by adding a prefix-based weighting that boosts similarity scores when strings share a common prefix. This makes it particularly suited for structured fields such as titles or author names, where early character agreement is especially informative.

\subsection{Baselines}
We compare \methodname against three publicly available rule-based libraries that rely on human-crafted heuristics or HTML DOM parsing: Trafilatura \citep{barbaresi-2021-trafilatura}, news-please \citep{Hamborg2017}, and Boilerpipe \citep{boilerpipe}. As model-based baselines, we include ReaderLM-v2 \citep{wang2025readerlmv2smalllanguagemodel}, Qwen2.5-1.5B, and Qwen2.5-32B-Instruct \citep{qwen2025qwen25technicalreport}.

A limitation of our evaluation is that the reference outputs are derived from the Fundus extraction rules, which introduces a potential bias in favor of \methodname, as it is trained to reproduce these very decisions. Other parsing libraries may follow different design philosophies regarding which content to include or exclude, and thus their outputs may diverge from the reference not due to lower quality, but due to differing extraction objectives. This should be taken into account when interpreting overlap-based metrics such as BLEU or ROUGE-L.

\section{Results}
\subsection{HTML-to-Plaintext}

\begin{table*}[!ht]
\centering
\begin{tabular}{lccccc}
\toprule
\textsc{Model} &
\textsc{BLEU} $\uparrow$ &
\textsc{METEOR} $\uparrow$ &
\textsc{ROUGE-L} $\uparrow$ &
\textsc{Lev.} $\downarrow$ &
\textsc{Jaro-W.} $\uparrow$ \\
\midrule
\textit{Pre-trained language models} \\
ReaderLM-v2 & 0.059 & 0.213 & 0.219 & 0.871 & 0.487 \\
Qwen2.5-32B-Instruct & 0.073 & 0.236 & 0.165 & 0.821 & 0.578 \\
\midrule
\textit{Fine-tuned language models} \\
Qwen2.5-1.5B & \underline{0.309} & \underline{0.466} & \underline{0.598} & \underline{0.448} & \underline{0.713} \\
Qwen2.5-1.5B + SimCTG & 0.276 & 0.434 & 0.552 & 0.508 & 0.690 \\
\methodname & \textbf{0.331} & \textbf{0.507} & \textbf{0.636} & \textbf{0.426} & \textbf{0.729} \\
\bottomrule
\end{tabular}
\caption{Results for HTML-to-JSON task.}
\label{tab:json-results}
\end{table*}

\Cref{tab:plaintext-results} shows the performance of all methods on the HTML-to-plaintext task. Among all fine-tuned approaches, \methodname achieves the strongest performance across all metrics, outperforming both rule-based parsing libraries and other model-based baselines.

Comparing against rule-based libraries, \methodname consistently outperforms both news-please and Boilerpipe across all reported metrics, demonstrating that task-specific fine-tuning on high-quality supervision can surpass hand-crafted heuristics for structured news extraction.

Among model-based approaches, we observe a clear performance gap between pre-trained and fine-tuned models. Pre-trained models, both ReaderLM-v2 and Qwen2.5-32B-Instruct, perform substantially worse despite their scale, indicating that zero-shot generalization is insufficient for accurate HTML-to-plaintext extraction. Fine-tuning drastically closes this gap, with even the smallest fine-tuned model, Qwen2.5-1.5B, achieving competitive performance with the rule-based libraries. Furthermore, the addition of SimCTG training yields consistent improvements over standard fine-tuning across all metrics, confirming the benefit of contrastive objectives for structured text generation. Finally, \methodname outperforms Qwen2.5-1.5B across all measures, suggesting that continued pretraining on HTML data, as provided by the ReaderLM backbone, further improves performance on structured news parsing beyond standard fine-tuning alone.

\begin{table*}[!htbp]
\centering
\begin{tabular}{lccc}
\toprule
\textsc{Model} &
\textsc{F1 (All)} &
\textsc{F1 (Valid JSON)} &
\textsc{Valid JSON} $\uparrow$ \\
\midrule
\multicolumn{4}{c}{\textit{Pre-trained language models}} \\
ReaderLM-v2 & 0.008 & 0.025 & 0.200 \\
Qwen2.5-32B-Instruct & 0.006 & 0.020 & 0.144 \\
\midrule
\multicolumn{4}{c}{\textit{Fine-tuned language models}} \\
Qwen2.5-1.5B & \textbf{0.595} & \textbf{0.786} & \underline{0.644} \\
Qwen2.5-1.5B + SimCTG & \underline{0.542} & 0.713 & \textbf{0.656} \\
\methodname & 0.523 & \underline{0.782} & 0.533 \\
\bottomrule
\end{tabular}
\caption{
JSON extraction performance. \textsc{F1 (All)} is computed over all outputs, counting invalid JSON as empty predictions. \textsc{F1 (Valid JSON)} is computed only on syntactically valid JSON outputs. \textsc{Valid JSON (\%)} denotes the fraction of generations that could be parsed as valid JSON.
}
\label{table:json-f1-merged}
\end{table*}

\subsection{HTML-to-JSON}

\Cref{tab:json-results} presents results on the HTML-to-JSON task. Consistent with the plaintext results, pre-trained models without fine-tuning perform poorly across all metrics, with BLEU, METEOR, and ROUGE-L scores below 0.24 and high Levenshtein distances for both ReaderLM-v2 and Qwen2.5-32B-Instruct. This confirms that zero-shot application of general-purpose language models is insufficient for reliable structured metadata extraction.

Task-specific fine-tuning again yields substantial improvements across all metrics. \methodname achieves the best overall performance on overlap-based metrics, outperforming all baselines on every reported measure. Notably, it improves over Qwen2.5-1.5B by +0.022 BLEU, +0.041 METEOR, and +0.038 ROUGE-L, while also reducing Levenshtein distance by 0.022. Interestingly, adding SimCTG training to Qwen2.5-1.5B does not yield the same gains as observed in the plaintext task, suggesting that the contrastive objective may be less beneficial for structured JSON generation.

To complement the overlap-based evaluation, we further assess syntactic validity and attribute-level accuracy in \Cref{table:json-f1-merged}, reporting F1 scores for key generation alongside the fraction of syntactically valid outputs. Among fine-tuned models, Qwen2.5-1.5B attains the highest F1 both over all outputs (0.595) and when restricted to valid JSON (0.786), while maintaining a valid JSON rate of 64.44\%. The SimCTG variant achieves the highest syntactic validity (65.56\%) but lower F1 scores, suggesting a trade-off between structural correctness and semantic accuracy. \methodname achieves a lower valid JSON rate (53.33\%) and F1 over all outputs (0.523), but its F1 restricted to valid outputs (0.782) is on par with Qwen2.5-1.5B, indicating that when \methodname does produce valid JSON, the extraction quality is competitive. Overall, these results confirm that task-specific fine-tuning is essential for effective HTML-to-JSON extraction, and that continued pretraining on HTML data supports competitive semantic accuracy even under stricter structural constraints.

Finally, rule-based parsing libraries are excluded from this evaluation, as they do not produce structured metadata in the format defined by Fundus.

\section{Ablations}

\begin{figure*}[ht]
    \includegraphics[width=\linewidth]{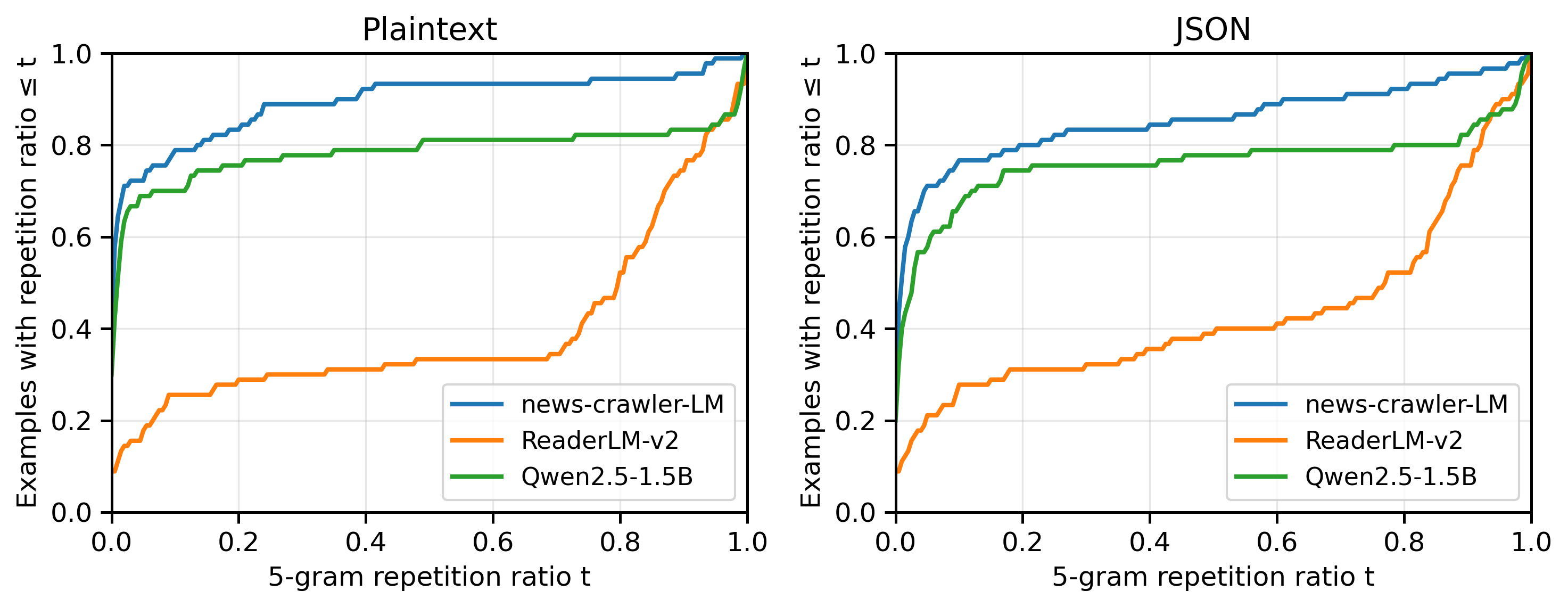}
    \caption{Cumulative distribution of 5-gram repetition ratios for plaintext (left) and JSON (right) generation. Each curve shows the fraction of outputs whose repetition ratio falls below threshold $t$. A curve shifted to the upper left indicates less repetitive generation. \methodname consistently produces the least repetitive outputs across both tasks.}
    \label{fig:repetition-distribution}
\end{figure*}

\subsection{Repetition and Degeneration Analysis}

Text degeneration in the form of repetitive outputs is a known failure mode of autoregressive language models \citep{holtzman2020curiouscaseneuraltext}. We analyze this tendency by measuring 5-gram repetition in the generated outputs, computing the fraction of outputs in which any 5-gram exceeds a repetition threshold of $t = 0.1$, i.e., appears in more than 10\% of tokens. A higher low-repetition rate thus indicates more fluent and diverse generation. Results are shown in \Cref{tab:low-repetition-rate}.

\begin{table}
\centering
\begin{tabular}{lc}
\toprule
\textsc{Model} & \textsc{Low-Rep. Rate} \\
\midrule
\textit{Pre-trained} \\
ReaderLM-v2 & 0.078 \\
\midrule
\textit{Fine-tuned} \\
Qwen2.5-1.5B & 0.667 \\
\methodname & \textbf{0.767} \\
\bottomrule
\end{tabular}
\caption{
Fraction of outputs with a 5-gram repetition ratio below $t=0.1$.
}
\label{tab:low-repetition-rate}
\end{table}

The pre-trained ReaderLM-v2 exhibits a very low low-repetition rate (0.078), indicating severe degeneration in the absence of task-specific fine-tuning. Among fine-tuned models, \methodname achieves the highest low-repetition rate (0.767), outperforming Qwen2.5-1.5B by 10 absolute points. This suggests that the SimCTG-based contrastive training objective, which explicitly encourages isotropic token representations, is effective at reducing repetitive generation in the context of structured HTML extraction.

To further characterize the distribution of repetitive outputs, \Cref{fig:repetition-distribution} plots the cumulative fraction of examples with a 5-gram repetition ratio below threshold $t$ for both tasks. Across both plaintext and JSON generation, \methodname consistently dominates the cumulative distribution, indicating that a larger fraction of its outputs exhibit low repetition at any given threshold. ReaderLM-v2 shows a markedly different profile, with its distribution rising slowly and remaining flat over a wide range of $t$, suggesting that a substantial portion of its outputs suffer from severe repetition. Qwen2.5-1.5B in between FundusCrawler and ReaderLM, performing comparably to \methodname at very low thresholds but falling behind as $t$ increases. These results are consistent across both tasks and corroborate the quantitative findings in \Cref{tab:low-repetition-rate}, confirming that the SimCTG-based training objective leads to more fluent and less degenerate outputs.

\subsection{Output Quality Distribution Analysis.}

\begin{figure*}[ht]
    \includegraphics[width=\linewidth]{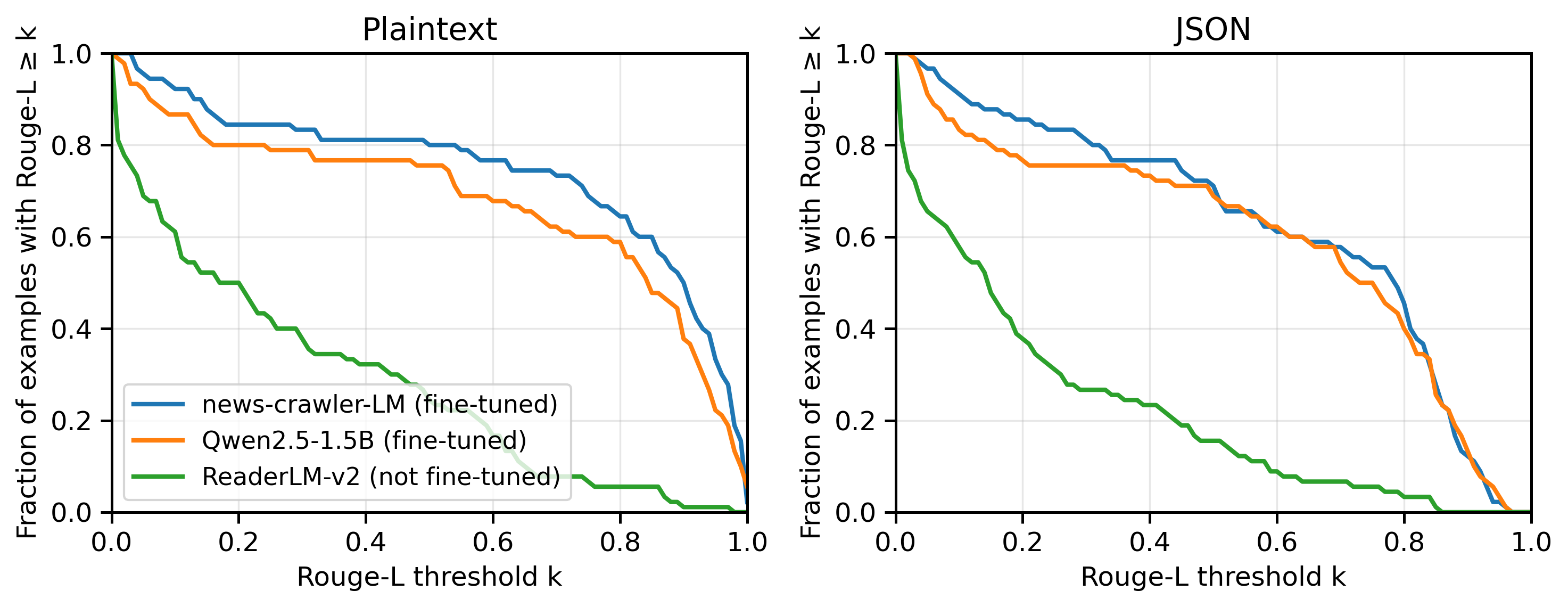}
    \caption{Complementary cumulative distribution of ROUGE-L scores for plaintext (left) and JSON (right) generation. Each curve shows the fraction of outputs achieving a ROUGE-L score of at least $k$. A curve shifted to the upper right indicates higher overall extraction quality. \methodname consistently achieves the highest ROUGE-L scores across both tasks.}
    \label{fig:rougel_distribution}
\end{figure*}

\Cref{fig:rougel_distribution} presents the complementary cumulative distribution of ROUGE-L scores across all test examples for both tasks. \methodname consistently dominates the distribution in both plaintext and JSON generation, meaning that a larger fraction of its outputs exceed any given ROUGE-L threshold $k$. For instance, in the plaintext task, \methodname maintains above 80\% of outputs with ROUGE-L $\geq 0.4$, whereas Qwen2.5-1.5B drops below this fraction considerably earlier. ReaderLM-v2, which is not fine-tuned, degrades rapidly and approaches zero well before $k = 1.0$, confirming that task-specific fine-tuning is critical for reliable extraction quality. Notably, the gap between \methodname and Qwen2.5-1.5B is more pronounced in the plaintext task than in JSON generation, where the two fine-tuned models perform more similarly across most of the distribution.

\section{Related Work}
Early work on web content extraction relied on heuristic and rule-based processing of HTML structure, often using DOM tree analysis to separate content from boilerplate \citep{10.1007/s11280-004-4873-3}. While effective for stable websites, these approaches degrade on heterogeneous layouts, motivating later learning-based methods that jointly model DOM structure and text to improve robustness and generalization \citep{Lin_2020, deng2022domlmlearninggeneralizablerepresentations}.

\noindent\textbf{Heuristic / Rule-Based Approaches.}~Recent line of works in forms of content extraction libraries transform web pages into text without relying on LLMs, typically combining heuristics and human-crafted rules to remove boilerplate content \citep{barbaresi-2021-trafilatura, Hamborg2017, pomikalek-etal-2012-building, boilerpipe, Finn2001FactOF, boilerplateremoval}. While broadly applicable, such generic extractors do not generalize well due to publisher-specific markup and achieve lower extraction precision. For example, Fundus \citep{dallabetta-etal-2024-fundus} relies on manually curated, publisher-specific rules, enabling high extraction accuracy but requiring substantial effort to adapt to new publishers. Our work builds on these developments by using rule-based extractions as low-noise supervision, enabling language models to generalize to new publishers.

\noindent\textbf{Machine Learning-Based Approaches.}~On the other hand, LLMs have been explored as universal HTML-to-text extractors, converting HTML to markup \citep{gur2023understandinghtmllargelanguage}. Prior work shows that explicitly modeling HTML structure and additional signals such as layout or visual cues can improve extraction performance \citep{Tan_2025, jung2022dontreadjustlook}, and LLMs have been applied to large-scale document preprocessing  \citep{xu2024cleanerpretrainingcorpuscuration} or interactive extraction settings \citep{bohra2025weblistsextractingstructuredinformation}. However, general-purpose LLMs often have billions of parameters, motivating our work on specialized long-context models for the HTML-to-text task \citep{kim2025nextevalevaluationtraditionalllm, wang2025readerlmv2smalllanguagemodel}.

\section{Conclusion}

We introduced \methodname, a small, domain-specialized language model that converts raw HTML from news websites into plaintext or structured JSON. Our approach utilizes rule-based extractions from the Fundus library as supervision signal, builds on the domain-adapted ReaderLM-v2 backbone, and incorporates an additional contrastive training objective to improve output stability and reduce degeneration.

Our evaluation shows that \methodname learns consistent parsing behavior and outperforms large general-purpose LLMs across all considered metrics, while requiring substantially fewer computational resources. The model also surpasses similarly sized language models fine-tuned on the same dataset, demonstrating the importance of the additional contrastive objective. Moreover, our quality distribution analysis indicates that the majority of outputs achieve high ROUGE-L scores, with only less than 20\% having a ROUGE-L scores $<0.6$.


Overall, our results show that targeted fine-tuning provides a practical alternative to manual rule engineering and computationally intensive large-scale LLM inference for web article extraction.

\section*{Limitations}
Although \methodname achieves strong performance, several limitations remain. First, the approach relies on high-quality rule-based supervision from Fundus, which may introduce biases or omissions present in the source extraction logic. While the models generalize to unseen publishers, they may still fail when confronted with radically different markup conventions, dynamically rendered content, or paywalled layouts.

Second, although contrastive learning reduces degeneration, a non-negligible portion of outputs still contain malformed JSON structures or missing fields, indicating the need for format-constrained decoding or post-hoc validation.

Third, the current work focuses primarily on English and German publishers, and generalization to low-resource languages and multilingual mixed-layout pages remains an open challenge. 

Fourth, inference on very long documents may be bottleneck using classical models using self-attention. In our observations, generating an extraction for a single long HTML document without specific inference optimizations can take several minutes. Although this remains more efficient (in terms of compute) than very large LLM-based approaches, it may not meet latency requirements in high-throughput production settings. Addressing these limitations requires further exploration of models using mixture-of-experts \citep{Cai_2025} of linear attention \citep{katharopoulos2020transformersrnnsfastautoregressive}, in which the computational complexity can be reduced.

Finally, as we train generative models to parse HTML into structured outputs, our approach is inherently susceptible to hallucinations. In some cases, the model may introduce information that is not explicitly supported by the input document, for example by inferring missing fields or generating plausible but incorrect metadata. While such behavior is common in sequence generation models, it poses particular challenges for information extraction tasks, where correctness and faithfulness to the source are critical. Mitigating hallucinations may require stronger input grounding, constrained decoding strategies, or explicit verification mechanisms.


\bibliography{references}

\end{document}